\documentclass[letterpaper, 10 pt, conference]{ieeeconf}  

\IEEEoverridecommandlockouts                              

\usepackage{graphicx}
\usepackage{multirow}
\usepackage{booktabs}
\usepackage{makecell}
\usepackage{arydshln}
\usepackage{xcolor}
\usepackage{placeins}
\usepackage{subcaption} 
\usepackage{hyperref}

\title{\LARGE \bf
ReVLA: Reverting Visual Domain Limitation\\of Robotic Foundation Models

}

\author{Sombit Dey$^{1}$, Jan-Nico Zaech$^{1}$, Nikolay Nikolov$^{1}$, Luc Van Gool$^{1}$, Danda Pani Paudel$^{1}$
\thanks{$^{1}$INSAIT, Sofia University “St. Kliment Ohridski”, Bulgaria}%
\thanks{{Corresponding Author: \tt\small jan-nico.zaech@insait.ai}}%
}

\begin{document}

\maketitle
\thispagestyle{empty}
\pagestyle{empty}

\begin{abstract}
Recent progress in large language models and access to large-scale robotic datasets has sparked a paradigm shift in robotics models transforming them into generalists able to adapt to various tasks, scenes, and robot modalities. A large step for the community are open Vision Language Action models which showcase strong performance in a wide variety of tasks. In this work, we study the visual generalization capabilities of three existing robotic foundation models, and propose a corresponding evaluation framework.

Our study shows that the existing models do not exhibit robustness to visual out-of-domain scenarios. This is potentially caused by limited variations in the training data and/or catastrophic forgetting, leading to domain limitations in the vision foundation models. We further explore OpenVLA, which uses two pre-trained vision foundation models and is, therefore, expected to generalize to out-of-domain experiments. However, we showcase catastrophic forgetting by DINO-v2 in OpenVLA through its failure to fulfill the task of depth regression.

To overcome the aforementioned issue of visual catastrophic forgetting, we propose a gradual backbone reversal approach founded on model merging. This enables OpenVLA -- which requires the adaptation of the visual backbones during initial training -- to regain its visual generalization ability. Regaining this capability enables our ReVLA model to improve over OpenVLA by a factor of 77\% and 66\% for grasping and lifting in visual OOD tasks.
 Comprehensive evaluations, episode rollouts and model weights are available on the \textcolor{blue}{\href{https://insait-institute.github.io/ReVLA/}{ReVLA Page}}

\FloatBarrier



\end{abstract}

\bstctlcite{IEEEexample:BSTcontrol}

\section{INTRODUCTION}
Generalist robot foundation models based on vision-language models (VLMs) offer a promising path toward developing dexterous policies that generalize across tasks, embodiments, and environments. This expectation is primarily based on the capabilities of the underlying large language models (LLMs) like Llama~\cite{touvron2023llamaopenefficientfoundation} that have been demonstrated in a range of real-world applications. VLMs equip LLMs with foundational vision encoders, such as DINO~\cite{oquab2024dinov2learningrobustvisual}, CLIP~\cite{radford2021clip} or SigLIP~\cite{zhai2023sigmoid}, enabling them to perform complex tasks that require perception and reasoning. Training of robot foundation models -- which output the robot's action trajectories -- requires training on large amounts of robot-specific data. Doing so adequately is a delicate task, which we study in this paper while focusing on the visual generalization capabilities in the out-of-domain (OOD) scenarios.

We begin by observing that most demonstrations used for training robotic foundation models are acquired in controlled lab settings and prioritize trajectory variability rather than comprehensive coverage of all possible interaction objects and environments. Simultaneously, training protocols such as those in OpenVLA~\cite{kim2024openvla} require adapting not only the LLM-based reasoning model but also the vision backbones to achieve high performance on in-domain tasks. Although pre-trained vision models themselves are known for generalizing well across diverse settings, their adaptation during the training of robotic foundation models can lead to visual domain limitations due to catastrophic forgetting.

\begin{figure}[t]
    \centering
    \includegraphics[width=1\linewidth]{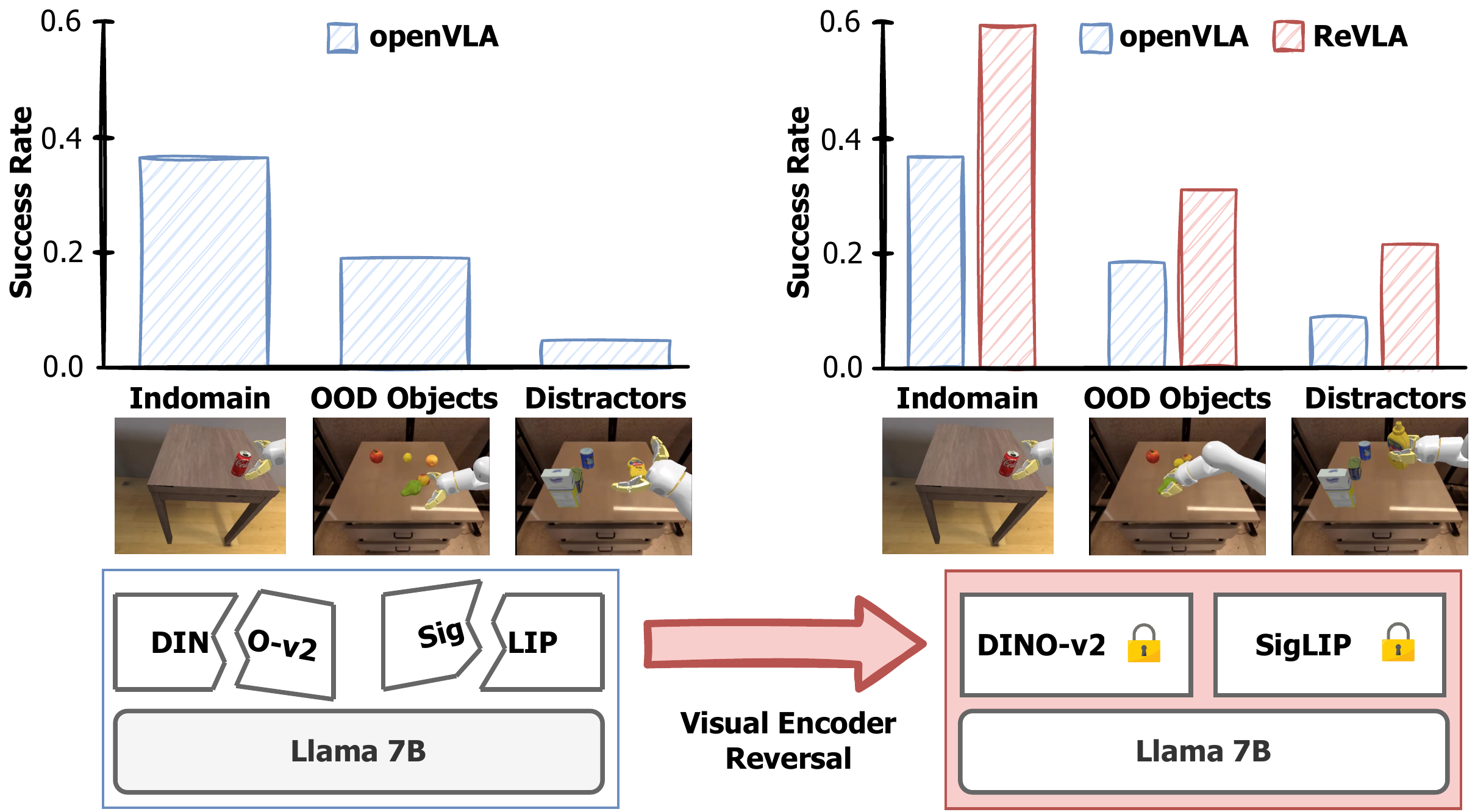}
    \caption{ The OpenVLA model tuned on fractal data struggles with out-of-domain objects and in the presence of distractors (left), due to catastrophic forgetting in its DINO-v2 and SigLIP vision encoders. Our ReVLA (DS gradual) model addresses this by reverting the vision encoders to their original pre-trained weights, leading to improved overall performance across all three settings (right).  
    }
    \label{fig:Teaser}
\end{figure}

To quantify the impact of this training approach and assess the overall generalization capability of robotic foundation models, we develop a realistic out-of-domain evaluation framework based on the SIMPLER~\cite{li24simpler} environment. This evaluation framework includes challenging visual tasks using OOD objects and visual distractors. We use it to evaluate the visual generalization performance of three recent open-source robotic foundation models: (1) RT-1~\cite{brohan2023rt1roboticstransformerrealworld}, which pioneered end-to-end generalist policies and employs an EfficientNet backbone alongside a transformer model for predicting low-level robot control commands; (2) Octo~\cite{octo_2023}, which is built on a diffusion policy integrated with a transformer model; and (3) OpenVLA~\cite{kim2024openvla}, the first publicly available vision-language-action model for robot control, built on the Prismatic~\cite{karamcheti2024prismatic} VLM that combines Llama~\cite{touvron2023llamaopenefficientfoundation} with DINO-v2 and SigLIP. In all three cases, the vision encoders are either trained or fine-tuned on visually limited robotics data.  

In our evaluations, we observe that all three models exhibit limited visual generalization performance. However, OpenVLA shows strong potential due to its powerful VLM foundation. Building on this, we propose a model reversal approach for the VLM's visual encoders, motivated by the expectation that the original pre-trained vision encoders should generalize well. Our approach builds on the concept of gradual model merging that is widely utilized for pretrained foundation models to slowly return to the pretrained weights. Overall, it enables a return to these generalizable visual embeddings, even when the training protocol for robotic foundation models requires fine-tuning the entire architecture, as is the case with OpenVLA.

In summary, we make three major contributions:
\begin{itemize}
    \item A framework for evaluating visual OOD performance of robotic foundation models containing OOD objects and distractors.
    \item A model reversal approach to regain generalization performance of visual encoders, even when the adaptation is required during training.
    \item An in-depth analysis of our ``ReVLA" approach through thorough experimental evaluations.
\end{itemize}

\section{Related Works}
\subsection{Vision Language Models}
VLMs~\cite{karamcheti2024prismatic, liu2023visualinstructiontuning, beyer2024paligemma} combine the strong generalization and few-shot learning capabilities of LLMs with the ability to understand the rich information present in image data. This is unlocked by utilizing one or more vision foundation models. In this process, the vision model serves the purpose of translating the spatially high-dimensional and strongly correlated information into a spatially sparse representation, but high-dimensional in the feature space, and thus, well suited for processing in an LLM.

Early steps were made by CLIP~\cite{radford2021clip} by embedding images and text into a joint feature space that is trained using contrastive learning on a large corpus of image-text pairs. While the model is not directly suitable for visual question answering, it provides powerful features for downstream applications like image classification and retrieval. BLIP~\cite{li2022blip} improves these capabilities by allowing interactions between text and vision inputs using the Q-former~\cite{zhang2023qformer} architecture.

By combining a pretrained Llama~\cite{touvron2023llamaopenefficientfoundation} LLM with a CLIP~\cite{radford2021clip} encoder as input to the LLM, LLaVA~\cite{liu2023visualinstructiontuning} utilizes the potential present in internet-scale text data for vision tasks. This paradigm is further used by approaches such as Flamingo~\cite{Alayrac2022FlamingoAV}, BLIP-2~\cite{2023LiBLIP2} or PaliGemma~\cite{beyer2024paligemma}. While architectures based on CLIP are well suited for image-level understanding, robotic problems require pixel-level semantic and spatial information. This is approached by combining image encoders in PixelLM~\cite{ren2023pixellm} and Prismatic~\cite{karamcheti2024prismatic}, which is the basis of OpenVLA~\cite{kim2024openvla} and our work.

\subsection{Generalist Robotics Models}

\begin{table}[tb]
\centering
\begin{tabular}{p{1.0cm} p{2.0cm} p{1.5cm} p{2.6cm}}
\textbf{Model}  & \textbf{Vision Backbone}      & \textbf{Reasoning Model}             & \textbf{Dataset} \\ \hline

RT-1            & EfficientNet+FiLM           & Transformer                     & Fractal \\ \hline
Octo            & Shallow CNN                   & Transformer+ Diffusion & OpenX Embodiment \\ \hline
OpenVLA         & DINOv2+SigLIP               & Llama V2                & OpenX Embodiment \\ \hline
\end{tabular}
\caption{Summary of generalist robotic foundation models.}
\label{tab:list_policies}
\end{table}

Similar to the development in VLMs, progress in robotics strongly revolves around the development of generalist models trained on large-scale datasets to perform diverse tasks across various embodiments and scenarios. RT1~\cite{brohan2023rt1roboticstransformerrealworld} introduced a language-conditioned model that directly maps from text and image inputs to robot control commands. By training a similar architecture on the OpenX dataset, which includes over 1 million trajectories from 22 robot embodiments, RT1-X~\cite{open_x_embodiment_rt_x_2023} demonstrated superior performance, transferring knowledge across diverse robots and tasks~\cite{open_x_embodiment_rt_x_2023}.

RT-2~\cite{zitkovich2023rt2} advanced this further by building on top of VLA models, including PaLI-X~\cite{Chen2023palix} and PaLM-E~\cite{driess2023palme}, co-fine-tuning them on both internet-scale and robotic data. This allowed RT-2 to exhibit first emergent reasoning abilities and superior generalization, outperforming RT1 in handling unseen objects and environments~\cite{zitkovich2023rt2}.

Octo~\cite{octo_2023} approaches robotic manipulation using diffusion-based by denoising the next actions performed by the robot. Its architecture supports both language and goal image conditioning and improves the generalization in manipulation tasks compared to RT1-X~\cite{octo_2023}.

The current state-of-the-art model, OpenVLA~\cite{kim2024openvla} is the first open-source generalist model utilizing a LLM. It is based on the prismatic VLM~\cite{karamcheti2024prismatic}, and trained on the OpenX embodiment dataset. It surpasses both specialist and foundational models, including RT-2, across various benchmarks. A summary of the publicly available current robotic foundation models is provided in Table~\ref{tab:list_policies}.

\subsection{Digital Twins and Photorealistic simulators}
Simulators are important tools in robotics research, offering a cost-effective and scalable alternative to real-world data collection and experiments~\cite{Muratore2022randomized_simulation}. Real-world robotics evaluation is strongly constrained by time and cost and does not offer reproducibility, especially across different setups. To overcome these challenges, simulation environments like SAPIEN~\cite{Xiang_2020_SAPIEN}, Habitat~\cite{szot2021habitat}, and Isaac Sim~\cite{makoviychuk2021isaac} provide nearly photorealistic proxies for real-world experiments.

In this work, we build on top of and extend standardized benchmarks for robotics, since they facilitate direct and reproducible comparisons across models. While a sim-to-real gap~\cite{zhao2020sim} also remains in current approaches, considerable progress has been made in matching domains and a strong correlation between simulation and real-world performance has been verified for foundation models in SIMPLER~\cite{li24simpler}, which serves as the basis of our work.

For building evaluation scenarios,  ALFRED~\cite{ALFRED20}, GraspNet~\cite{fang2020graspnet}, and MotionBenchMaker~\cite{chamzas2021motionbenchmaker} provide comprehensive task definitions to evaluate perception and manipulation. In our work, we utilize the grasping task definition in SIMPLER~\cite{li24simpler} and extend it with the YCB object dataset~\cite{Calli2015_ycb} to model OOD objects and distractor settings.

\subsection{Model Merging}
Efficiently reusing models trained on different datasets is crucial for applications that utilize large-scale data, since training all models from scratch is prohibitively costly and requires large amounts of energy. Model merging~\cite{ainsworth2023gitrebasinmergingmodels,yang2024model} is an approach to gain the different capabilities of pretrained models trained on different tasks and datasets, by combining their parameter weights. The approach has been shown to help save training GPU hours, mitigate catastrophic forgetting~\cite{alexandrov2024mitigating}, and boost performance in domain generalization~\cite{arpit2022ensembleaveragesimprovingmodel}.
In VLMs, model merging~\cite{sung2023empirical} results in better out-of-domain generalization and better performance and mitigates the inference overhead from using multiple encoders. In this work, we utilize linear model merging to revert domain-limited encoders to well-generalizing models.

\section{Background}
This section provides the background information relevant to our model reversal approach. As the foundation of our model, we introduce Prismatic~\cite{karamcheti2024prismatic} and OpenVLA~\cite{kim2024openvla} as a corresponding VLM and VLA pair. For our benchmark, we utilize the SAPIEN~\cite{Xiang_2020_SAPIEN} simulator with the SIMPLER~\cite{li24simpler} environment which is introduced in the second part.

\subsection{Vision Language Action Models}
OpenVLA \cite{kim2024openvla} is currently the only open-source, generalist, foundational robotics model for manipulation tasks. OpenVLA builds upon the Prismatic VLM \cite{karamcheti2024prismatic} architecture and weights, by tuning the model on the OpenX dataset\cite{open_x_embodiment_rt_x_2023} to output discrete action tokens for end-effector control similar to RT1 and RT2~\cite{brohan2023rt1roboticstransformerrealworld,zitkovich2023rt2}.

The model first processes the visual input using SigLIP~\cite{zhai2023sigmoid} and DINO-v2~\cite{oquab2024dinov2learningrobustvisual} to generate features respectively. The DINO-v2~\cite{oquab2024dinov2learningrobustvisual}  encoder is a self-supervised trained vision encoder that allows for spatial and relational understanding, while SigLIP~\cite{zhai2023sigmoid}, trained using contrastive learning between image-language pairs, is incorporated for semantic and linguistic multimodal understanding. A tokenizer projects the image features to a Llama 6.7B\cite{touvron2023llamaopenefficientfoundation} LLM that processes both, language instructions and the tokenized visual features.

During training on the OpenX dataset end effector control is obtained by seven output action tokens ($ \Delta x, \Delta y, \Delta z, \Delta \alpha, \Delta \beta , \Delta \gamma, \Delta$gripper) for every image-language pair as input. For this task, the 256 least used tokens are overwritten. During training of OpenVLA on the OpenX dataset~\cite{open_x_embodiment_rt_x_2023} tuning of the whole model, including the two vision foundation models is required to achieve sufficiently stable robot trajectories~\cite{kim2024openvla}.

\subsection{Simpler-Benchmark}
Our benchmark extends the SIMPLER~\cite{li24simpler} environment, which itself is an adaptation of the Maniskill benchmark \cite{mu2021maniskill}, based on the SAPIEN Simulator \cite{Xiang_2020_SAPIEN}. We chose SIMPLER as it addresses the two main challenges in simulated robotics evaluations, the control and visual gap. Therefore, results achieved in SIMPLER closely match real-world comparisons between algorithms and allow us to avoid hardly reproducible robot experiments.

\textbf{Control Gap:} The goal of reducing the control gap between simulated and real-world environments is to ensure that policy actions performed in simulation produce similar outcomes on the robot's end-effector when applied to the real robot, thus minimizing the real-to-sim gap. 
SIMPLER~\cite{li24simpler} aims to mitigate these differences by optimizing the parameters of the robot controller in the simulator such that the simulator trajectories, when rolled out in an open-loop manner for a given sequence of actions, are close to the real-world trajectories.

\textbf{Vision Gap:} Discrepancies in the visual appearance between a simulator and real-world observation cause a domain shift, which can harm the behavior of learned policies~\cite{Zhao2020SimtoReal, Chebotar2019closingSimtoReal}. To mitigate the visual gap, SIMPLER proposes a “green screening” approach. This involves overlaying the real-world environment's background with the images rendered by the simulator, providing a realistic visual representation.

Thus, SIMPLER is a realistic setting to evaluate robot foundation models in simulations that are reproducible, low-cost, and fast, and have demonstrated to preserve the rank correlation in model performances to real robot experiments.

\begin{figure}[tb]
    \centering
    \begin{subfigure}[b]{0.48\linewidth}
        \centering
        \includegraphics[width=\textwidth]{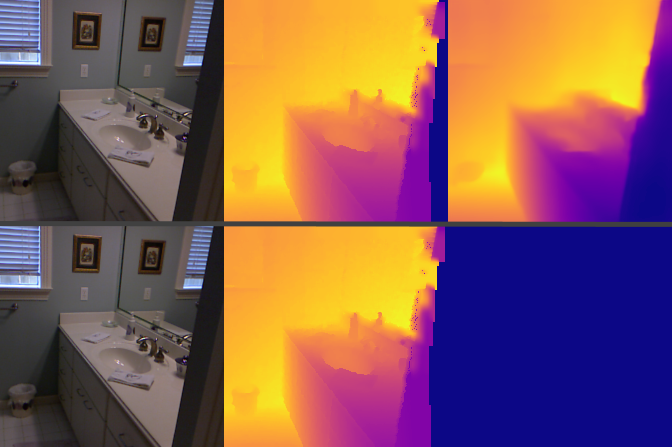} 
        \caption{DPT.}
        \label{fig:dpt}
    \end{subfigure}
    \hfill
    \begin{subfigure}[b]{0.48\linewidth}
        \centering
        \includegraphics[width=\textwidth]{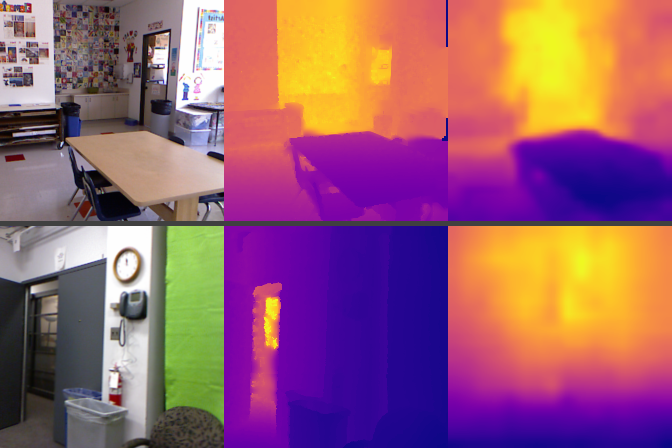} 
        \caption{Linear.}
        \label{fig:linear}
    \end{subfigure}
    \caption{Depth regression from original DINO-v2 (top-right) and OpenVLA DINO-v2 (bottom-right) using DPT head~(a) and linear probing (b). Each row: input image, ground-truth, predicted depth. The original DINO-v2 estimates depths correctly, whereas OpenVLA DINO-v2 performs poorly due to catastrophic forgetting during OpenVLA training. }
    \label{fig:depth}
\end{figure}

\section{Regaining Vision Encoder Capabilities}
Our method aims at reverting the vision foundation models used in OpenVLA back to their original state in Prismatic, pretrained on a large corpus of diverse vision-language tasks. Therefore, the following section first uncovers the catastrophic forgetting present in the OpenVLA backbones, based on this, two strategies for model reversal are proposed. Finally, baselines for a finetuned version of OpenVLA are established to allow us for a fair comparison between our ReVLA and the base OpenVLA models.

\subsection{Catastrophic Forgetting in OpenVLA}
The OpenX dataset~\cite{open_x_embodiment_rt_x_2023} is an important basis for the training of robotic foundation models, containing more than 1 million episodes of robotic demonstrations. This diversity spans a wide range of robot embodiments, tasks, and trajectories, sufficient for training generalist models that have the ability to operate across various modalities. However, the variation present in OpenX dataset is primarily focused on task complexity and embodiment rather than on visual and scene diversity.

In contrast, large-scale VLMs are trained on datasets that primarily emphasize visual diversity and thus are substantially richer in this aspect. Even more, vision foundation models like DINO-V2~\cite{beyer2024paligemma}, are trained on partly private data which further enhances diversity and generalizability.

This lack of visual diversity presents a particular challenge when training robotic foundation models. RT-2~\cite{zitkovich2023rt2} approaches this challenge by co-training on the WebLI dataset~\cite{Chen2023palix}, however, this also considerably adds to the training requirements, which are a limiting factor for open models like OpenVLA. The risk of the model forgetting is further amplified in OpenVLA due to the need for fine-tuning both the vision encoders and the LLM simultaneously to reach maximum performance, which is in contrast to the approach taken in the underlying Prismatic architecture.

To demonstrate the catastrophic forgetting in the visual encoders we utilize DINO-v2 features for depth estimation using linear probing as well as a dense prediction (DPT) transformer head~\cite{Ranftl2020dpt} as shown in Figure~\ref{fig:depth}. When tasked with regressing to depth information, the trained vision encoder in OpenVLA collapses to a constant output for DPT and generates depth maps without details for linear probing, reflecting a loss of spatial understanding, while the pretrained DINO-v2 features generate high-quality depth maps, especially with DPT. This collapse is a clear indicator of catastrophic forgetting, particularly within the vision backbone, as the model fails to retain core spatial competencies learned in earlier stages of training.

This evidence highlights a key challenge in training large-scale robotics models: although fine-tuning the entire architecture, including the vision backbone, can improve stability, it may reduce the model's ability to generalize, particularly in visually diverse environments.

\subsection{ReVLA: Reverting the Vision Backbones}
\label{sec:revla}
Approaching the challenge of catastrophic forgetting in VLA vision encoders, we propose ReVLA as an approach to revert back to the pretrained vision encoders and their inherent capabilities, while maintaining the training protocol of OpenVLA ensuring high-quality robot trajectories. Towards solving this task, we evaluate two different training approaches inspired by model merging as well as the influence of the two different encoders, SigLIP\cite{zhai2023sigmoid} and DINO-v2~\cite{oquab2024dinov2learningrobustvisual}. We base our work on the most powerful version of OpenVLA trained on the complete OpenX dataset~\cite{open_x_embodiment_rt_x_2023}, and perform the encoder reversal using the Fractal RT-1 robot action dataset~\cite{brohan2023rt1roboticstransformerrealworld}, since it is part of the OpenX dataset and contains a wide range of robotic tasks and uses the same embodiment as our simulator.

\textbf{OpenVLA}: We utilize OpenVLA trained on the full OpenX dataset without additional finetuning as our first baseline since it has shown good generalization performance across embodiments and basic emergent behavior.

\textbf{OpenVLA-fractal}: To account for a potential bias by finetuning on a well-fitting subset of the whole OpenX dataset, we finetune the original OpenVLA model on the Fractal {RT-1} robot action dataset which becomes our second baseline model. We follow the original OpenVLA training protocol and tune the whole model including the visual encoders for the same number of steps and the same hyperparameters to have a fair comparison with our approach and evaluation.

\textbf{ReVLA}: ReVLA follows the OpenVLA architecture and is initialized with weights trained on OpenX. As the core component of ReVLA, visual encoder reversal is performed during training. The reversal process is inspired by continuous model merging approaches, where the weights of one model are gradually integrated into a second differently trained one. However, different from model merging, our process ends with the fully pretrained weights, rather than with an interpolation, as the pretrained model covers a considerably larger domain that is required for OOD tasks and emergent behavior.

We follow a linear merging schedule of the vision encoder weights, which slowly reverts from the OpenVLA weights $\theta_{OpenVLA}$ to the pretrained encoder weights, according to
\begin{equation}
    F_{theta} = (1 - \alpha) \theta_{OpenVLA} + \alpha \theta_{Pretrained},
    \label{eq:model_merging}
\end{equation}
where $\alpha$ is the weight of the mixing value of pre-trained dino weights. For computational efficiency, the linear transition is implemented as a step-wise curriculum of changing $\alpha$, every $n$ steps by a value $\frac{1}{k}$. The total training steps is $N = k \times n$. In this work, we choose $N$ = 100K steps and $n$ =10k steps.

OpenVLA utilizes two visual encoders, DINO-V2 for spatial and 3D reasoning tasks and SigLIP for semantic understanding. In our process, both models are reverted: DINO-v2 since OOD generalization requires strong spatial understanding beyond in-domain tasks, and SigLIP, since interactions in an open world require the capability to describe objects that are not present in the robotic training data. Furthermore, while training ReVLA we keep the vision backbone frozen. We refer to this method as ReVLA (DS gradual) in the following.
To verify our design choices, we define four models:

1) ReVLA (DS gradual), gradually reverts both backbones.

2) ReVLA (DS flip), does not employ a gradual model-merging approach but restores the weights of both backbones at the start of training.

3) ReVLA (D gradual), where only the DINO-v2 encoder is slowly reverted and SigLIP is kept frozen.

4) ReVLA (D flip), which combines the two approaches and only reverts DINO-v2 using hard flipping at the start of the reversal process.

\section{Experiments and Results}
\label{sec:Exp}
Our experiments investigated the following questions, 
\begin{itemize}
    \item How does the out-of-domain performance of ReVLA differ from OpenVLA?
    \item How does reverting vision backbones affect indomain performance?
    \item What is the best protocol to revert a vision backbone for a VLA?
\end{itemize}   
Given the computational cost of training on the entire OpenX dataset, we limit our experiments to the Fractal dataset~\cite{brohan2023rt1roboticstransformerrealworld}, which consists of approximately 73.5k robot trajectories. This dataset is based on Table-Top and kitchen-top scenes used in the Google robot manipulation tasks and includes template-based language annotations (e.g., “Pick up the Coke Can”). All experiments were conducted using one Nvidia A100 node, training each model for 100,000 steps with identical hyperparameters.

\subsection{Embodied OOD Benchmark}

\textbf{In Domain experiments:}
To evaluate the overall impact of reverting the vision encoders of openVLA, we first assess ReVLA on the SIMPLER benchmark~\cite{li24simpler}, which represents in-domain scenarios for openVLA. We chose SIMPLER as it minimizes the sim-to-real gap by using simulation environments closely aligned with the Fractal dataset.

The SIMPLER benchmark comprises two sub-evaluations: Visual Matching and the Variant Aggregation protocols, which take two different approaches for closing the domain gap; The \textit{Visual Matching} protocol simulates photorealistic conditions as closely as possible, ensuring minimal visual discrepancy with the Fractal dataset. The \textit{Variant Aggregation} protocol uses strong domain randomization, altering factors like lighting, table textures, backgrounds, and visual distractors to assess the model’s robustness to environmental changes. While Variant Aggregation represents diverse scenarios it visually does not closely resemble a real scenario. Thus, as ReVLA is strongly aimed at visual performance, we chose visual matching as the basis for our evaluation.

In Table~\ref{tab:SimplerEval}, we compare ReVLA to other foundational models, including RT-1-X, Octo, and OpenVLA. To avoid bias we choose our baseline methods as described in Section~\ref{sec:revla} and train on the Fractal dataset for an equal number of steps. We follow the notion of SIMPLER, where for an episode to be successful, the robot has to pick up the target object and lift it up for a duration of time. 

\textbf{Out of domain experiments:}
To evaluate the robustness of our models to OOD tasks, we modify the simple \textit{Pick Coke Can} task by introducing new, unseen objects into the scene. These experiments are conducted following the \textit{Visual Matching} protocol and are designed to assess the models' ability to generalize in more realistic, complex environments. For each setting we evaluate over 36 episodes, resulting in a total of 216 OOD evaluation scenarios.

\textbf{OOD objects:}
In the first set of experiments, we introduce several OOD objects, selected from the YCB dataset~\cite{Calli2015_ycb} (\textit{Pear}, \textit{Mustard Bottle}, and \textit{Tomato Can}) to test the generalization capability of models to unseen objects manipulations. Each object presents a distinct set of difficulties:
\begin{itemize}
\item \textit{Tomato Can}: This object closely resembles in-domain items such as the \textit{Coke Can} and \textit{RedBull Can}, providing a test of the open-set recognition ability as well as the ability to differentiate between visually similar objects.
\item \textit{Pear}: In addition to not being present in the training dataset, the asymmetrical shape of the pear adds complexity to the task of grasping, requiring the model to demonstrate awareness of the 3D shape and adapt its manipulation strategy accordingly.
\item \textit{Mustard Bottle}: As a significantly larger object compared to the typical items in the dataset, the Mustard Bottle evaluates the model’s ability to adapt its policy for size variations and to execute robust grasping behaviors in such cases.
\end{itemize}

\textbf{OOD Objects with Distractor}:
To further challenge the models, we introduce additional distractors around the target objects. Picking a single object in isolation does not fully assess visual robustness. By surrounding the target object with other items, we aim to evaluate the models' visual grounding capabilities, as the correct object must be selected based on the language prompts and visual clues.

These distractors are chosen from the YCB dataset~\cite{Calli2015_ycb}, as well as from the objects available in SIMPLER. Thus, this experiment can further evaluate a bias of the model to grasp objects that were present during training.
Finally, distractors often occlude the target object, making the task even more demanding by forcing the model to work under partial visibility.

By combining OOD objects with distractors, these experiments highlight the model’s ability to handle complex scenes, demonstrating both visual robustness and the capacity for multimodal reasoning under challenging conditions.

\subsection{Results}
\begin{table}[tb]
    \centering
    \begin{tabular}{p{0.2cm} p{2.1cm} p{0.9cm} p{0.9cm} p{0.9cm} p{0.9cm} }
        \multirow{3}{*}{\makecell{}} & \multirow{2}{*}{\makecell{Policy}} & \multicolumn{4}{c}{Pick Coke Can}\\
        \cmidrule(lr){3-6} 
         & & Horizontal & Vertical & Standing & Average \\
        
        \midrule
        \multirow{6}{*}{\rotatebox[origin=c]{90}{\makecell{\centering Var Aggregation~~}}} & RT-1  & 0.969 & 0.760 & 0.964 & 0.897 \\
         & RT-1-X & 0.569 &0.204 &0.698 &0.490  \\
        \cdashline{2-6}
        
         & Octo-Base & 0.005& 0.000 &0.013 &0.006  \\
         & OpenVLA & \textbf{0.644} & 0.218 & \textbf{0.729} & \textbf{0.530} \\
        \cdashline{2-6}
         & OpenVLA-fractal & 0.483 & 0.231 & 0.368 & 0.361 \\
         & ReVLA (flip) & 0.360 & 0.422 & 0.469 & 0.417 \\
         & ReVLA (gradual) & 0.483 & \textbf{0.469} & 0.610 & 0.521  \\
         
        \midrule
        \multirow{6}{*}{\rotatebox[origin=c]{90}{\makecell{\centering Visual Matching~~}}} & RT-1  & 0.969 & 0.900 & 0.710 & 0.857 \\
         & RT-1-X &0.860 & 0.790 & 0.480 & 0.710  \\
        \cdashline{2-6}
            
         & Octo-Base & 0.210 &0.210 &0.090 &0.170  \\
         & OpenVLA & 0.310 & 0.030 & 0.190 & 0.177  \\
        \cdashline{2-6}  
         & OpenVLA-fractal &  \textbf{0.655}& 0.135 & 0.295 & 0.361 \\
         & ReVLA (flip) & 0.5475 & 0.357 & 0.726 & 0.544  \\
         & ReVLA (gradual) & 0.600 & \textbf{0.413} & \textbf{0.792} & \textbf{0.600} \\
    \end{tabular}
    \caption{In domain evaluation with SIMPLER~\cite{li24simpler}.}
    \label{tab:SimplerEval}
\end{table}
\begin{table*}[tb]
    \centering
    \begin{tabular}{p{2.8cm} p{1.cm} p{1.cm} p{1.cm} p{1.cm} p{1.cm} p{1.cm} p{1.cm} p{1.cm} p{1.cm}}
        \multirow{2}{*}{\makecell{Policy}} & \multicolumn{3}{c}{Single} & \multicolumn{3}{c}{Distractor} & \multicolumn{3}{c}{Overall} \\
        \cmidrule(lr){2-4} \cmidrule(lr){5-7} \cmidrule(lr){8-10}
         & Pear & Mustard Bottle & Tomato Can & Pear & Mustard Bottle & Tomato Can & Single & Distractors & Total \\
        \midrule
        RT1-X & 0.222 & 0.000 & 0.118 & 0.167 & 0.000 & 0.059 & 0.113 & 0.075& 0.094 \\
        Octo & 0.000 & 0.000 & 0.000 & 0.000 & 0.000 & 0.000 & 0.000 & 0.000& 0.000 \\
        OpenVLA  & 0.194 & 0.083 & 0.389 & 0.056 & 0.028 & 0.222 & 0.222 & 0.102 & 0.162 \\
        OpenVLA-fractal & 0.139 & 0.028 & 0.389 & 0.056 & 0.028 & 0.167 & 0.185 & 0.084 & 0.135 \\
        \midrule
        ReVLA (D flip)  & 0.306  & 0.056 & 0.306 & 0.222 & 0.000 & 0.389 & 0.223 & 0.204 & 0.213 \\
        ReVLA (DS flip) & \textbf{0.440} & \textbf{0.194} & 0.361 & \textbf{0.500} & 0.083 & 0.139 & 0.330 & \textbf{0.241} & \textbf{0.287} \\
        ReVLA (D gradual) & 0.222 & 0.139 & \textbf{0.556} & 0.167 & 0.000 & \textbf{0.472} & 0.306 & 0.213 & 0.259 \\
        ReVLA (DS gradual) & 0.389 & 0.110 & 0.528 & 0.306 & \textbf{0.110} & 0.220 & \textbf{0.340} & 0.213 & 0.278 \\
    \end{tabular}
    \caption{Success rate of robotics foundation models in OOD tasks.}
    \label{tab:SimplerEvalOOD}
\end{table*}

We evaluate ReVLA using both SIMPLER protocols in Table~\ref{tab:SimplerEval}, variant aggregation as well as visual matching for their in-domain performance. 
Using visual matching, our models outperform the OpenVLA baselines by 42.3\% and 23.9\% for the OpenX and Fractal tuned models respectively.
This demonstrates that ReVLA retains and improves upon OpenVLA's capabilities for in-domain tasks.

Using variant aggregation, the performance for a wider range of tasks parameters is evaluated. In this scenario, our best model outperforms OpenVLA trained on Fractal by 14.9\% and is comparable with the OpenX-trained model with a drop of 0.8\% in performances. 
This may be explained by the smaller variety of tasks present in Fractal.

Our OOD evaluation is presented in Table~\ref{tab:SimplerEvalOOD}. Compared to the in-domain task of visual matching, all models lose performance, which is the expected behavior due to the domain gap. Nevertheless, all ReVLA models improve the OOD performance compared to OpenVLA, with the best OOD model ReVLA (DS flip) improving by 12.5\%, which corresponds to a factor of 77\%. The strongest loss of performance is encountered for RT1, which is the best model in the in-domain visual matching task, and Octo, which is not also to solve any OOD task.

\subsection{Ablation}
\begin{table}
    \centering
    \begin{tabular}{l|cc}
         Model & Success Lift & Success Grasp \\
          \midrule
          openVLA & 0.162 & 0.348 \\ 
          openVLA-fractal & 0.135 & 0.317 \\
          ReVLA (Flip) & \textbf{0.287} & 0.495\\
          ReVLA (Gradual) & 0.278 & \textbf{0.579} \\
    \end{tabular}
    \caption{Grasping and lifting success in OOD tasks.}
    \label{tab:partial_success}
\vspace{-5mm}
\end{table}

We conducted an ablation study to assess the design choices in our ReVLA architecture, specifically focusing on two key aspects: (1) the impact of a gradual transition during the reversal process, and (2) whether the reversal should be applied only to the DINO-v2 encoder or to both visual encoders (DINO-v2 and SigLIP).

The performance results, shown in Table~\ref{tab:SimplerEvalOOD}, indicate that ReVLA~(Flip) marginally outperforms ReVLA~(Gradual) by 0.9\%. However, upon further analysis of robot trajectories in Table~\ref{tab:partial_success}, which evaluates partial success (i.e., how often the robot successfully grasps the target object), ReVLA~(Gradual) demonstrates a clear advantage. It is considerably more effective in locating and grasping the correct target object compared to ReVLA~(Flip).

Notably, in many of the partially successful episodes, although the robot grasps the object, it either fails to lift it properly or the object slips shortly after being lifted. This suggests that the drop in overall performance, despite higher grasping success, may be attributed to a lack of precise control during the lifting phase.

Overall, we find the following answers to above questions,
\begin{itemize}
    \item Reverting both vision encoders, helps ReVLA to regain the zero-shot multi-modal capability of SigLIP and the representative power of DINO-v2, resulting in better performance in OOD scenarios.
    \item Reverting the vision encoders further helps the in-domain performance compared to openVLA, likely owing to a better and more precise spatial understanding of the Dino-V2 vision encoders.
    \item Our ablation study on OOD Task shows that both hard and gradual flipping of the vision encoder outperform OpenVLA, with gradual flipping leading to a more robust model.  
\end{itemize}

\begin{figure}
    \centering
    \includegraphics[width=\linewidth]{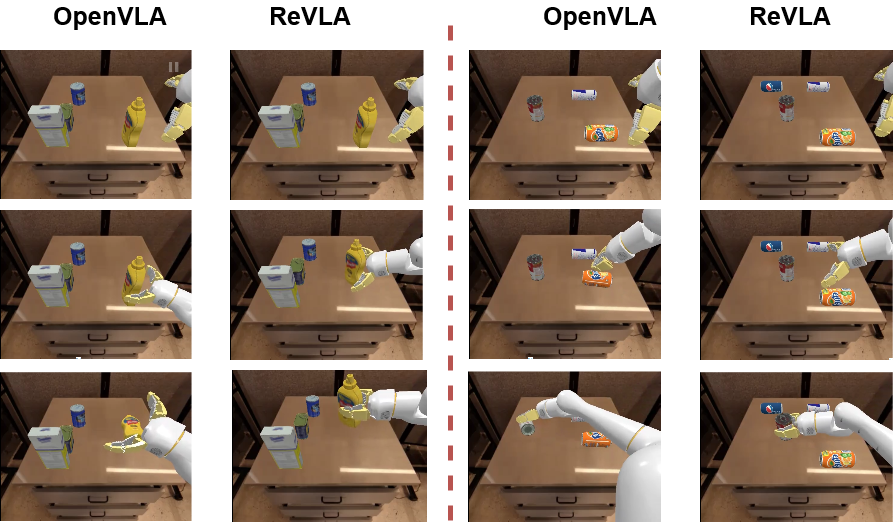}
    \caption{Left: Openvla and ReVLA are able to identify the target object. Openvla fails to grasp it. Right: OpenVLA tries to grasp the \textit{Fanta} can, which is nearest to the arm. ReVLA initially hesitates but eventually grasps the tomato can. }
    \label{fig:Qualitative}
\vspace{-5mm}
\end{figure}
\section{Conclusion and Future Work}
In this work, we evaluated and addressed the visual domain limitation in robotic generalist foundation models, specifically focusing on catastrophic forgetting in vision encoders when fine-tuning them on robotic datasets. We proposed ReVLA, a novel approach that reverts visual backbones to their pretrained states for models that require training of the whole architecture. This reversal approach successfully mitigated the catastrophic forgetting present in OpenVLA, restoring the visual generalization capabilities.

Despite these improvements, limitations remain. Our training was conducted on the Fractal dataset, which may not fully represent the diverse real-world scenarios in robotics, especially across embodiments. Additionally, an exploration of model merging techniques has the potential to yield even greater performance gains by integrating additional data from robotic datasets.

\addtolength{\textheight}{0cm}   



%
%
\section*{ACKNOWLEDGMENT}
\noindent This research was partially funded by the Ministry of Education and Science of Bulgaria (support for INSAIT, part of the Bulgarian National Roadmap for Research Infrastructure).



\pagebreak
\bibliographystyle{./IEEEtran}
\bibliography{./IEEEabrv,./references}

\end{document}